\documentclass[final]{cvpr}

\usepackage{times}
\usepackage{epsfig}
\usepackage{graphicx}
\usepackage{amsmath}
\usepackage{amssymb}
\usepackage{comment}

\usepackage[pagebackref=true,breaklinks=true,colorlinks,bookmarks=false]{hyperref}

\def\cvprPaperID{10} 
\def\confYear{CVPR 2021}

\begin{document}

\title{Sign Language Production: A Review}

\author{Razieh Rastgoo$^{1,2}$, Kourosh Kiani$^2$,Sergio Escalera$^3$, Mohammad Sabokrou$^1$  \\
$^1$Institute for Research in Fundamental Sciences (IPM)\\~~$^2$Semnan University~~ $^3$Universitat de Barcelona and Computer Vision
Center
\\

}

\maketitle
\begin{abstract}
 Sign Language is the dominant yet non-primary form of communication language used in the deaf and hearing-impaired community. To make an easy and mutual communication between the hearing-impaired and the hearing communities, building a robust system capable of translating the spoken language into sign language and vice versa is fundamental. To this end, sign language recognition and production are two necessary parts for making such a two-way system. Sign language recognition and production need to cope with some critical challenges. In this survey, we review recent advances in Sign Language Production (SLP) and related areas using deep learning. This survey aims to briefly summarize recent achievements in  SLP, discussing their advantages, limitations, and future directions of research.
\end{abstract}
\section{Introduction}
Sign Language is the dominant yet non-primary form of the communication language used in large groups of people in society. According to the World Health Organization (WHO) report in 2020, there are more than 466 million deaf people in the world \cite{who1}. There are different forms of sign languages employed by different nationalities such as USA \cite{asl}, Argentina \cite{argentina}, Poland \cite{poland}, Germany \cite{Germany}, Greek \cite{Greek}, Spain \cite{spain}, China \cite{china}, Korea \cite{Korean}, Iran \cite{Iran}, and so on. To make an easy and mutual communication between the hearing-impaired and the hearing communities, building a robust system capable of translating the spoken languages into sign languages and vice versa is fundamental. To this end, sign language recognition and production are two necessary parts for making such a two-way system. While the first part, sign language recognition, has rapidly advanced in recent years \cite{Rastgoo-rbm,  Rastgoo-multiview, Rastgoo-video, Rastgoo-pose-aware, Rastgoo-svd, RastgooSurvey}, the latest one, Sign Language Production (SLP), is still a very challenging problem involving an interpretation between visual and linguistic information \cite{stoll2020}. Proposed systems in sign language recognition generally map signs into the spoken language in the form of text transcription \cite{RastgooSurvey}. However, SLP systems perform the reverse procedure.\\ 
Sign language recognition and production are coping with some critical challenges \cite{RastgooSurvey,stoll2020}. One of them is the visual variability of signs, which is affected by hand-shape, palm orientation, movement, location, facial expressions, and other non-hand signals. These differences in sign appearance produce a large intra-class variability and low inter-class variability. This makes it hard to provide a robust and universal system capable of recognizing different sign types. Another challenge is developing a photo-realistic SLP system to generate the corresponding sign digit, word, or sentence from a text or voice in spoken language in a real-world situation. The challenge corresponding to the grammatical rules and linguistic structures of the sign language is another critical challenge in this area. Translating between spoken and sign language is a complex problem. This is not a simple mapping problem from text$/$voice to signs word-by-word. This challenge comes from the differences between the tokenization and ordering of words in the spoken and sign languages.\\ Another challenge is related to the application area. Most of the applications in sign language focus on sign language recognition such as robotics \cite{Dawes}, human–computer interaction \cite{Bachmann}, education \cite{Darabkh}, computer games \cite{Roccetti}, recognition of children with autism \cite{cai}, automatic sign-language interpretation \cite{yang}, decision support for medical diagnosis of motor skills disorders \cite{Butt}, home-based rehabilitation \cite{Cohen} \cite{Morando}, and virtual reality \cite{Vaitkev}. This is due to the misunderstanding of the hearing people thinking that deaf people are much more comfortable with reading spoken language; therefore, it is not necessary to translate the reading spoken language into sign language. This is not true since there is no guarantee that a deaf person is familiar with the reading and writing forms of a speaking language. In some languages, these two forms are completely different from each other. 
While there are some detailed and well-presented reviews in sign language recognition \cite{Ghanem,RastgooSurvey}, SLP suffers from such a detailed review. Here, we present a survey, including recent works in the SLP, with the aim of discussing advances and weaknesses of this area. We focus on deep learning-based models to analyze state-of-the-art on SLP.\\ 
The remainder of this paper is organized as follows. Section 2 presents a taxonomy that summarizes the main concepts related to SLP. Finally, section 3 discusses the developments, advantages, and limitations in SLP and comments on possible lines for future research.
\section{SLP Taxonomy}
\begin{figure}[t]
\begin{center}
\includegraphics[width=1\linewidth]{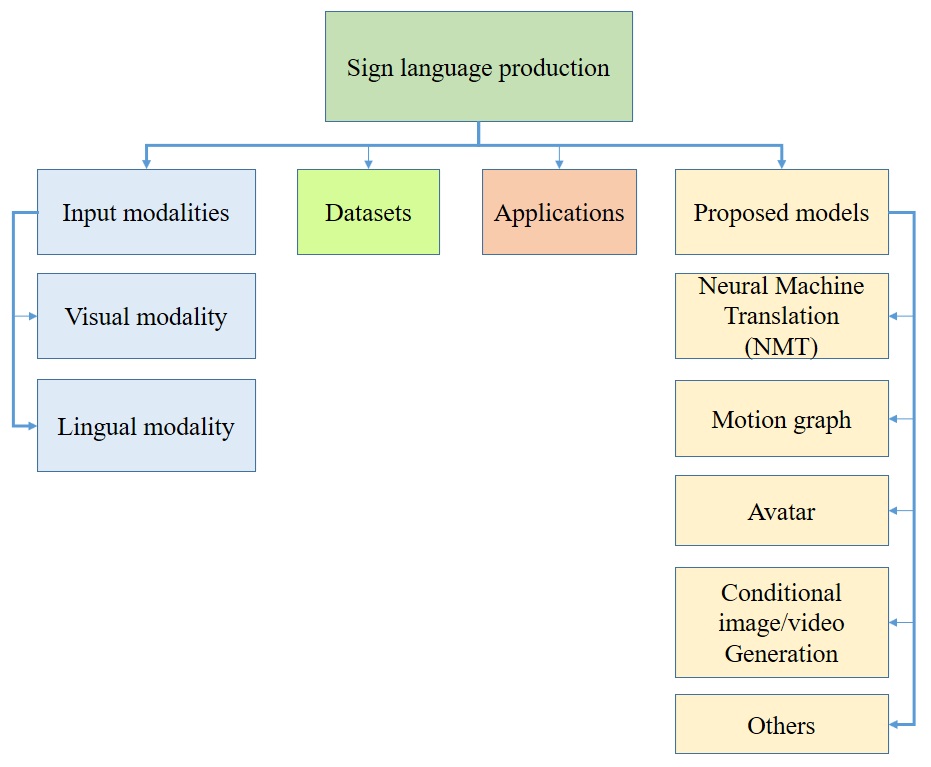}   
\end{center}
   \caption{The proposed taxonomy of the reviewed works in SLP.}
\label{fig:long}
\label{Fig 1}
\end{figure}

In this section, we present a taxonomy that summarizes the main concepts related to deep learning in SLP. We categorize recent works in SLP providing separate discussion in each category. In the rest of this section, we explain different input modalities, datasets, applications, and proposed models. Figure \ref{Fig 1} shows the proposed taxonomy described in this section.
\subsection{Input modalities}
Generally, vision and language are two input modalities in SLP. While the visual modality includes the captured image$/$video data, the linguistic modality contains the text input from the natural language. Computer vision and natural language processing techniques are necessary to process these input modalities.\\
\textbf{Visual modality:} RGB and skeleton are two common types of input data used in SLP models. While RGB images$/$videos contain high-resolution content, skeleton inputs decrease the input dimension necessary to feed to the model and assist in making a low-complex and fast model. Only one letter or digit is included in an RGB image input. The spatial features corresponding to the input image can be extracted using computer vision-based techniques, especially deep learning-based models. In recent years, Convolutional Neural Networks (CNN) achieved outstanding performance for spatial feature extraction from an input image \cite{Rastgoo-cnn}. Furthermore, generative models, such as Generative Adversarial Networks (GAN),
can use the CNN as an encoder or decoder block to generate a sign image$/$video. Due to the temporal dimension of RGB video inputs, the processing of this input modality is more complicated than the RGB image input. Most of the proposed models in SLP use the RGB video as input \cite{Camgoz,SaundersBMVC,Saunders,stoll2020}. An RGB sign video can corresponding to one sign word or some concatenated sign words, in the form of a sign sentence. GAN and LSTM are the most used deep learning-based models in SLP for static and dynamic visual modalities. While successful results have been achieved using these models, more effort is necessary to generate more lifelike sign images$/$videos in order to improve the communication interface with the Deaf community.\\
\textbf{Lingual modality:} Text input is the most common form of linguistic modality. To process the input text, different models are used \cite{See,Sutskever}. While text processing is low-complex compared to image$/$video processing, text translation tasks are complex. Among the deep learning-based models, the Neural Machine Translation (NMT) model is the most used model for input text processing. Other Seq2Seq models, such as Recurrent Neural Network (RNN)-based models, proved their effectiveness in many tasks. While successful results were achieved using these models, more effort is necessary to overcome the existing challenges in the translation task. One of the challenges in translation is related to domain adaptation due to different words styles, translations, and meaning in different languages. Thus, a critical requirement of developing machine translation systems is to target a specific domain. Transfer learning, training the translation system in a general domain followed by fine-tuning on in-domain data for a few epochs is a common approach in coping with this challenge. Another challenge is regarding the amount of training data. Since a main property of deep learning-based models is the mutual relation between the amount of data and model performance, large amount of data is necessary to provide a good generalization. Another challenge is the poor performance of machine translation systems on uncommon and unseen words. To cope with these words, byte-pair encoding, such as stemming or compound-splitting, can be used for rare words translation. As another challenge, the machine translation systems are not properly able to translate long sentences. However, the attention model partially deals with this challenge for short sentences. Furthermore, the challenge regarding the word alignment is more critical in the reverse translation, that is translating back from the target language to the source language.
\subsection{Datasets}
\begin{figure}[t]
\begin{center}
\includegraphics[width=1\linewidth]{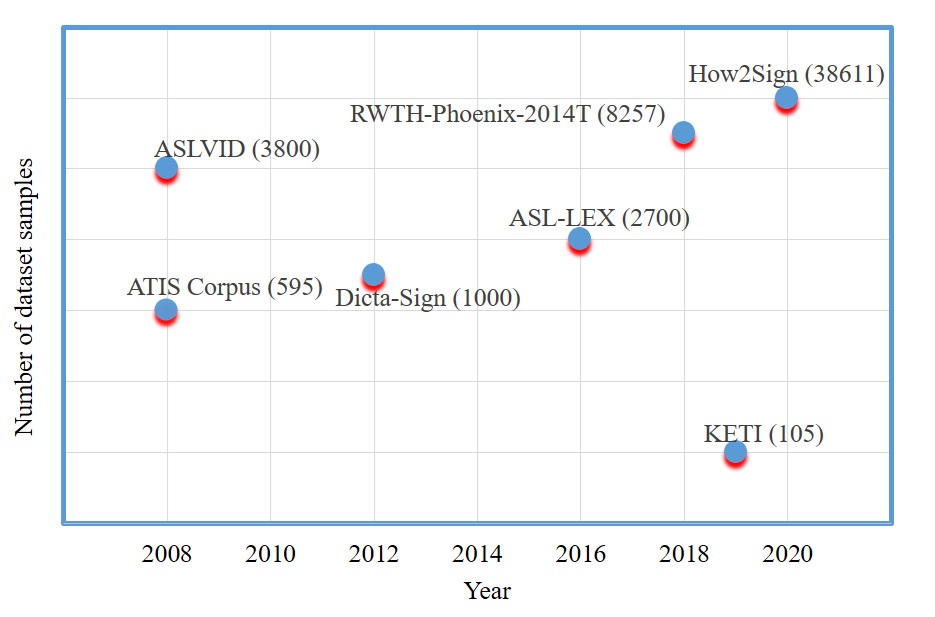}   
\end{center}
   \caption{SLP datasets in time. The number of samples for each dataset is shown in brackets}
\label{fig:long}
\label{Fig 3}
\end{figure}

While there are some large-scale and annotated datasets available for sign language recognition, there are only few publicly available large-scale datasets for SLP. Two public datasets, RWTH-Phoenix-2014T \cite{Necati2018} and How2Sign \cite{Duarte} are the most used datasets in sign language translation. The former includes German sign language sentences that can be used for text-to-sign language translation. This dataset is an extended version of the continuous sign language recognition dataset, PHOENIX-2014 \cite{Forster}. RWTH-PHOENIX-Weather 2014T includes a total of 8257 sequences performed by 9 signers. There are 1066 sign glosses and 2887 spoken language vocabularies in this dataset. Furthermore, the gloss annotations corresponding to the spoken language sentences have been included in the dataset. The later dataset, How2Sing, is a recently proposed multi-modal dataset used for speech-to-sign language translation. This dataset contains a total of 38611 sequences and 4k vocabularies performed by 10 signers. Like the former dataset, the annotation for sign glosses have been included in this dataset.\\
Though RWTH-PHOENIX-Weather 2014T and How2Sign provided SLP evaluation benchmarks, they are not enough for generalization of SLP models. Furthermore, these datasets just include German and American sentences. In line with the aim of providing an easy to use application for mutual communication between the Deaf and hearing communities, new large-scale datasets with enough variety and diversity in different sign languages is required. The point is that the signs are generally dexterous and the signing procedure involves different channels, including arms, hands, body, gaze, and facial expressions simultaneously. To capture such gestures requires a trade-off between capture cost, measurement (space and time) accuracy, and the production spontaneity. Furthermore, different equipment is used for data recording such as wired Cybergloves, Polhemus magnetic sensors, headset equipped with an infrared camera, emitting diodes and reflectors. Synchronization between different channels captured by the aforementioned devices is key in data collection and annotation. Another challenge is related to the capturing complexity of the hand movement using some capturing devices, such as Cybergloves. Hard calibration and deviation during data recording are some difficulties of these acquisition devices. The synchronization of external devices, hand modeling accuracy, data loss, noise in the capturing process, facial expression processing, gaze direction, and data annotation are additional challenges. Given these challenges, providing a large and diverse dataset for SLP, including spoken language and sign language annotations, is difficult. Figure \ref{Fig 3} shows existing datasets for SLP.

\subsection{Applications}
With the advent of the potent methodologies and techniques in recent years, machine translation applications have become more efficient and trustworthy. One of the early efforts on machine translation is dated back to the sixties, where a model was proposed to translate from Russian to English. This model defined the machine translation task as a phase of encryption and decryption. Nowadays, the standard machine translation models fall into three main categories: rule-based grammatical models, statistical models, and example-based models. Deep learning-based models, such as Seq2Seq and NMT models, fall into the third category, and showed promising results in SLP.\\
To translate from a source language to a target language, a corpus to perform some preprocessing steps is needed, including boundary detection, word tokenization, and chunking. While there are different corpora for most spoken languages, sign language lacks from such a large and diverse corpora. American Sign Language (ASL), as the largest sign language community in the World, is the most-used sign language in the developed applications for SLP. Since Deaf people may not read or write the spoken language, they need some tools for communication with other people in society. Furthermore, many interesting and useful applications in Internet are not accessible for the Deaf community. The aforementioned projects aim to develop such tools. However, we are still far from having applications accessible for Deaf people with large vocabularies$/$sentences from real-world scenarios. One of the main challenges for these applications is a license right for usage. Only some of these applications are freely available. Another challenge is the lack of generalization of current applications, which are developed for the requirements of very specific application scenarios.

\subsection{Proposed models}
In this section, we review recent works in SLP. These works are presented and discussed in five categories: Avatar approaches, NMT approaches, Motion Graph (MG) approaches, Conditional image/video Generation approaches, and other approaches. Table \ref{Table 2 } presents a summary of the reviewed models in SLP. 
\subsubsection{Avatar Approaches}
In order to reduce the communication barriers between hearing and hearing-impaired people, sign language interpreters are used as an effective yet costly solution. To inform deaf people quickly in cases where there is no interpreter on hand, researchers are working on novel approaches to providing the content. One of these approaches is sign avatars. Avatar is a technique to display the signed conversation in the absence of the videos corresponding to a human signer. To this end, 3D animated models are employed, which can be stored more efficiently compared to videos. The movements of the fingers, hands, facial gestures, and body can be generated using the avatar. This technique can be programmed to be used in different sign languages. With the advent of computer graphics in recent years, computers and smartphones can generate high-quality animations with smooth transitions between the signs. To capture the motion data of deaf people, some special cameras and sensors are used. Furthermore, a computing method uses to be considered to transfer the body movements into the sign avatar \cite{Kipp-Avatar}.\\
Two ways to derive the sign avatars include the motion capture data and parametrized glosses. In recent years, some works have been developed exploring avatars animated from the parametrized glosses. VisiCast \cite{Bangham}, Tessa \cite{Cox}, eSign \cite{Zwitserlood}, dicta-sign \cite{Efthimiou}, JASigning \cite{VHG}, and WebSign \cite{Jemni} are some of them. These works need the sign video annotated via the transcription language, such as HamNoSys\cite{Prillwitz} or SigML \cite{Kennaway}. Although, the non-popularity of these avatars made them unfavorable in the deaf community. Under-articulated, unnatural movements, and missing non-manuals information, such as eye gaze and facial expressions, are some challenges of the avatar approaches. These challenges lead to misunderstanding the final sign language sequences. Furthermore, due to the uncanny valley, the reviewers do not feel comfortable \cite{Mori} with the robotic motion of the avatars. To tackle these problems, recent works focus on annotating non-manual information such as face, body, and facial expression \cite{Ebling-2015,EblingJohn-2013}.\\ 
Using the data collected from motion capture, avatars can be more usable and acceptable for reviewers (such as the Sign3D project by MocapLab \cite{Gibet}). Highly realistic results are achieved by avatars, but the results are restricted to a small set of phrases. This comes from the cost of the data collection and annotation. Furthermore, avatar data is not a scalable solution and needs expert knowledge to perform a sanity check on the generated data. To cope with these problems and improve performance, deep learning-based models, as the latest machine translation developments, are used. Generative models along with some graphical techniques, such as Motion Graph, are being recently employed \cite{stoll2020}.
\subsubsection{NMT approaches}
Machine translators are a practical methodology for translating from one language to another. The first translator comes back to the sixties where the Russian language was translated into English \cite{Hutchins}. The translation task requires preprocessing of the source language, including sentence boundary detection, word tokenization, and chunking. These preprocessing tasks are challenging, especially in sign language. Translating from a spoken language into sign language is a non-trivial task. The ordering and the number of glosses do not necessary match the words of the spoken language sentences.\\
Nowadays, there are different types of machine translators, mainly based on grammatical rules, statistics, and examples \cite{Othman}. As an example-based methodology, some research works have been developed by focusing on translating from text into sign language using Artificial Neural Networks (ANNs), namely NMT \cite{Bahdanau}. NMT uses ANNs to predict the likelihood of a word sequence, typically modeling entire sentences in a single integrated model.\\ 
To enhance the translation performance of long sequences, Bahdanau et al. \cite{Bahdanau} presented an effective attention mechanism. This mechanism was later improved by Luong et al. \cite{Luong}. Camgoz et al. proposed a combination of a seq2seq model with a CNN to translate sign videos to spoken language sentences \cite{Camgoz-2018}. Guo et al. \cite{Guo-2018} designed a hybrid model including the combination of a 3D Convolutional Neural Network (3DCNN) and Long Short Term Memory (LSTM)-based encoder-decoder to translate from sign videos to text outputs. Results on their own dataset show a 0.071 \%\ improvement margin of the precision metric compared to state-of-the-art models. Dilated convolutions and Transformer are two approaches also used for sign language translation \cite{Kalchbrenner,Vaswani}. Stoll et al. \cite{stoll2020} proposed a hybrid model to automatic SLP using NMT, GANs, and motion generation. The proposed model generates sign videos from spoken language sentences with a minimal level of data annotation for training. This model first translates spoken language sentences into sign pose sequences. Then, a generative model is used to generate plausible sign language video sequences. Results on the PHOENIX14T Sign Language Translation dataset show comparable results compared to state-of-the-art alternatives.\\
While NMT-based methods achieved successful results in translation tasks, some major challenges need to be solved. Domain adaptation is the first challenge in this area. Since the translation between different domains is affected by different rules, domain adaptation is a crucial requirement in developing machine translation systems targeted to a specific use case. The second challenge is regarding the amount of training data. Especially in deep learning-based models, increasing the amount of data leads to better results. Another difficulty is dealing with uncommon words. The translation models perform poorly on these words. Words alignment and adjusting the beam search parameters are other challenges for NMT-based models. The promising results of current deep learning-based models set an underpin to future research in this area.
\subsubsection{Motion Graph approaches}
Motion Graph (MG), as a computer graphic method for dynamically animating characters, is defined as a directed graph constructed from motion capture data. MG can generate new sequences to satisfy specific goals. In SLP, MG can be combined with NMT-based network to make a continuous-text-to-pose translation. One of the early efforts on MG is dated back to 2002, where a general framework was proposed by Kovar et al. \cite{Kovar} for extracting particular graph walks that satisfy a user’s specifications. Distance between two frames was defined as a distance between two point clouds. To make the transitions, alignment and interpolation of the motions and positions were used between the joints. Finally, the branch and bound search algorithm was applied to the graph. 
In another work, Arikan and Forsyth \cite{Arikan} used the joint positions, velocities, and accelerations parameters to define the distance between two consequence frames. In addition, the discontinuity between two clips was calculated using a smoothing function. After summarizing the graph, the random search was applied to the graph. Two-layer representation of motion data in another approach was proposed by Lee et al. \cite{Lee}. In the first layer, data was modeled as a first-order Markov process and the transition probabilities were calculated using the distances of weighted joint angles and velocities. A cluster analysis was performed on the second layer, namely cluster forest, to generalize the motions. Stoll et al. \cite{stoll2020} proposed an MG for continuous SLP using pose data. The sign glosses were embedded to an MG with the transition probabilities provided by an NMT decoder at each time step.\\ 
Although MG can generate plausible and controllable motion through a database of motion capture, it faces some challenges. The first challenge is regarding the access to data. To show the model potential with a truly diverse set of actions, a large set of data is necessary.  Scalability and computational complexity of the graph to select the best transitions are the other challenges in MG. Furthermore, since the number of edges leaving a node increases with the size of the graph, the branching factor in the search algorithm will increase as well. 
\subsubsection{Conditional image/video generation}
The field of automatic image/video generation has experienced a remarkable evolution in recent years. However, the task of video generation is challenging since the content between consequence frames has to be consistent, showing a plausible motion. These challenges are more difficult in SLP due to the need for human video generation. The complexity and variety of actions and appearances in these videos are high and challenging. Controlling the content of the generated videos is crucial yet difficult.\\
With the recent advances in deep learning, the field of automatic image/video generation has seen different approaches employing neural network-based architectures, such as CNNs \cite{Chen,Oord}, RNNs \cite{Gregor,Oord-pixelRNN}, Variational Auto-Encoders (VAEs) \cite{VAE}, conditional VAEs \cite{CVAE}, and GAN \cite{GAN}. VAEs and GANs are generally combined to benefit from the VAE’s stability and the GAN’s discriminative nature. Most relevant to SLP, a hybrid model, including a VAE and GAN combination, has been proposed to image generation of people \cite{Ma-2017,Siarohin} and video generation of people performing sign language \cite{Stoll-sign,Vasani}. Furthermore, there are some models for image/video generation that can be used in SLP. For example, Chen and Koltun \cite{Chen} proposed a CNN-based model to generate photographic images given semantic label maps. Van den Oord et al. \cite{Oord} proposed a deep learning-based model, namely PixelRNNs, to sequentially generate the image pixels along the two spatial dimensions. Gregor et al. \cite{Gregor} developed an RNN-based architecture, including an encoder and a decoder network to compress the real images presented during training and refine images after receiving codes. Karras et al. \cite{Karras} designed a deep generative model, entitled StyleGAN, to adjust the image style at each convolution layer. Kataoka et al. \cite{Kataoka} proposed a model using the combination of GAN and attention mechanism. Benefiting from the attention mechanism, this model can generate images containing high detailed content.\\
While deep learning-based generative models have recently achieved remarkable results, there exist major challenges in their training. Mode collapse, non-convergence and instability, suitable objective function, and optimization algorithm are some of these challenges. However, several strategies have been recently proposed to address a better design and optimization of them. Appropriate design of network architecture, proper objective functions, and optimization algorithms are some of the proposed techniques to improve the performance of deep learning-based models.
\subsubsection{Other models}
In addition to the previous categories, some models have been proposed to SLP using different deep learning models. For example, Saunders et al. \cite{Saunders} proposed a Progressive Transformers, as a deep learning-based model, to generate continuous sign sequences from spoken language sentences. In another work, Zelinka and Kanis \cite{Zelinka} designed a sign language synthesis system focusing on skeletal model production. A feed-forward transformer and a recurrent transformer, as deep learning-based models, along with the attention mechanism have been used to enhance the model performance. Saunders et al. \cite{Saunders-arxiv} proposed a generative-based model to generate photo-realistic continuous sign videos from text inputs. They combined a transformer with a Mixture Density Network (MDN) to manage the translation from text to skeletal pose. Tornay et al. \cite{Tornay} designed a SLP assessment approach using multi-channel information (hand shape, hand movement, mouthing, facial expression). In this approach, two linguistic aspects are considered: the generated lexeme and the generated forms. \\
Using the capabilities of different methods in this category has led to successful results in SLP. However, the challenge of the model complexity still remains an open issue. Making a trade-off between accuracy vs. task complexity is a key element.  

\begin{table*}[h!]
\thispagestyle{empty}
\caption{\label{Table 2 } Summary of deep SLP models.}
\begin{center}
{\small
 \noindent\begin{tabular}{p{0.8cm}p{0.7cm}p{1.5cm}p{2.2cm}p{1.8cm}p{7cm}}
 \hline \hline
 \textbf{Year} & \textbf{Ref} & \textbf{Feature} & \textbf{Input modality} &  \textbf{Dataset} & \textbf{Description}\\
 \hline \hline
 2011 & \cite{Kipp-Avatar} & Avatar & RGB video & ViSiCAST & \textbf{Pros.} Proposing a gloss-based tool focusing on the animation content evaluating using a new metric for comparing avatars with human signers.
\textbf{Cons.} Need to include non-manual features of human signers.\\
2016 & \cite{McDonald} & Avatar & RGB video & Own dataset & \textbf{Pros.}  Automatically adding realism to the generated images, low computational complexity.
\textbf{Cons.} Need to place the position of the shoulder and torso extension on the position of the avatar’s elbow, rather than the IK end-effector.\\
2016 & \cite{Gibet} & Avatar & RGB video & Own dataset & \textbf{Pros.} Easy to understand with high viewer acceptance of the sign avatars.
\textbf{Cons.} Limited to the small set of sign phrases.\\
2018 & \cite{Camgoz-2018} & NMT	& RGB video	& PHOENIX-Weather 2014T	& \textbf{Pros.} Robust to jointly align, recognize, and translate sign videos.
\textbf{Cons.} Need to align the signs in the spatial domain.\\
2018 & \cite{Guo-2018} & NMT & RGB video & Own dataset & \textbf{Pros.} Robust to align the word order corresponding to visual content in sentences.
\textbf{Cons.} Need to generalize to additional datasets.\\
2020 & \cite{stoll2020} & NMT, MG & Text & PHOENIX14T & \textbf{Pros.} Robust to minimal gloss and skeletal level annotations for model training. \textbf{Cons.} Model complexity is high.\\
2020 & \cite{Saunders}	& Others & Text	& PHOENIX14 & \textbf{Pros.} Robust to the dynamic length of output sign sequence.
\textbf{Cons.} Model performance can be improved including non-manual information.\\
2020 & \cite{Zelinka} & Others	& Text	& Czech news & \textbf{Pros.} Robust to the missing skeletons parts.
\textbf{Cons.} Model performance can be improved including information of facial expressions.\\
2020 & \cite{Saunders-arxiv} & Others & Text & PHOENIX14T & \textbf{Pros.} Robust to non-manual feature production.
\textbf{Cons.} Need to increase the realism of the generated signs.\\
2020 & \cite{Camgoz} & Others & Text	& PHOENIX14T &	\textbf{Pros.} No need to the gloss information.
\textbf{Cons.} Model complexity is high.\\
2020 & \cite{Saunders-BMVC} & Others & Text & PHOENIX14T & \textbf{Pros.} Robust to manual feature production. \textbf{Cons.} Need to increase the realism of the generated signs.\\
\hline
\noalign{\smallskip}
\end{tabular}
 }
\end{center}
\end{table*}
\section{Discussion}
\begin{figure}[t]
\begin{center}
\includegraphics[width=1\linewidth]{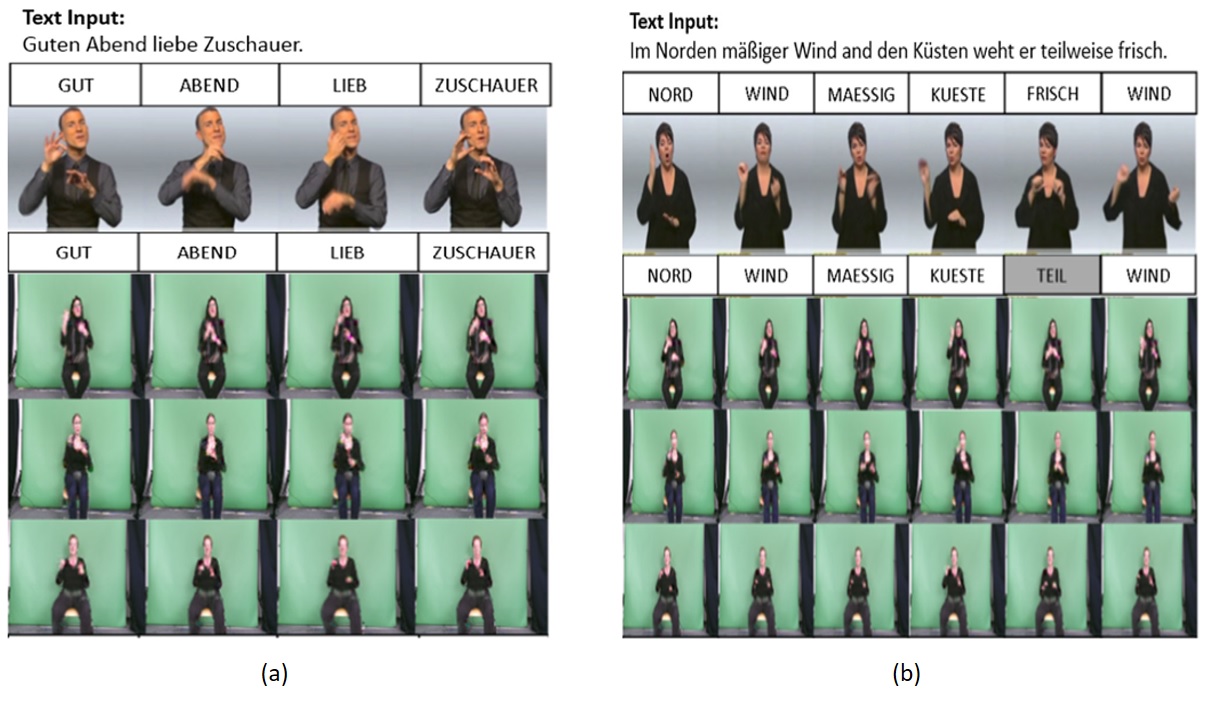}   
\end{center}
   \caption{Translation results from \cite{stoll2020}: (a) “Guten Abend liebe Zuschauer”. (Good evening dear viewers), (b) “Im Norden maessiger Wind an den Kuesten weht er teilweise frisch”. (Mild winds in the north, at the coast it blows fresh in parts).  Top row: Ground truth gloss and video, Bottom row: Generated gloss and video. This model combines an NMT network and GAN for SLP.}
\label{Fig 5}
\end{figure}

\begin{figure}[t]
\begin{center}
\includegraphics[width=1\linewidth]{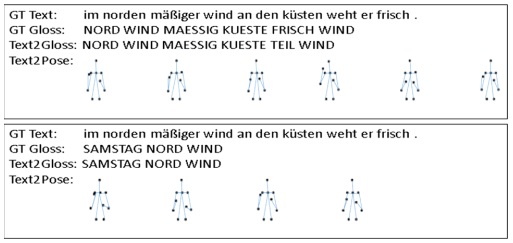}   
\end{center}
   \caption{Translation results from \cite{stoll2020}: Text from spoken language is translated to human pose sequences.}
\label{Fig 6}
\end{figure}

\begin{figure}[t]
\begin{center}
\includegraphics[width=1\linewidth]{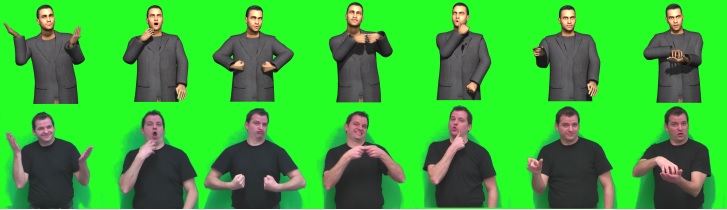}   
\end{center}
   \caption{Translation results from \cite{Kipp-Avatar}: A signing avatar is created using a character animation system. Top row: signing avatar, Bottom row: original video.}
\label{Fig 7}
\end{figure}

In this survey, we presented a detailed review on the recent advancements in SLP. We presented a taxonomy that summarizes the main concepts related to SLP. We categorized recent works in SLP providing separate discussion in each category. The proposed taxonomy covered different input modalities, datasets, applications, and proposed models. Here, we summarize main findings:\\
\textbf{Input modalities:} Generally, vision and language modalities are two input modalities in SLP. While the visual modality includes the captured image/video data, the linguistic modality contains the text input from natural language. Computer vision and natural language processing techniques are necessary to process these input modalities. Both categories benefit from deep learning approaches to improve model performance. RGB and skeleton are two common types of visual input data used in SLP models. While RGB images/videos contain high-resolution content, skeleton inputs decrease the parameter complexity of the model and assist in making a low-complex and fast model. GAN and LSTM are the most used deep learning-based models in SLP for visual inputs. While successful results were achieved using these models, more effort is necessary to generate more lifelike and high-resolution sign images/videos acceptable by the Deaf community. Among the deep learning-based models for lingual modality, the NMT model is the most used model for input text processing. Other Seq2Seq models, such as RNN-based models, proved their effectiveness in many tasks. While accurate results were achieved using these models, more effort is necessary to overcome the existing challenges in the translation task, such as domain adaptation, uncommon words, words alignment, and word tokenization.\\
\textbf{Datasets:} The lack of a large annotated dataset is one of the major challenges in SLP. The collection and annotation of sign language data is an expensive task that needs the collaboration of linguistic experts and native speakers. While there are some publicly available datasets for SLP \cite{Athitsos,Bungeroth,Necati2018,Caselli,Duarte,Ko,Matthes}, they suffer from weakly annotated data for sign language. Furthermore, most of the available datasets in SLP contain a restricted domain of the vocabularies/sentences. To make a real-world communication between the Deaf and hearing communities, access to a large-scale continuous sign language dataset, segmented on the sentence level, is necessary. In such dataset, a paired form of the continuous sign language sentence and the corresponding spoken language sentence needs to be included. Just a few datasets meet these criteria \cite{Camgoz-2018,Duarte,Ko,Zelinka} . The point is that most of the aforementioned datasets cannot be used for end-to-end translation  \cite{Camgoz-2018,Ko,Zelinka}. Two public datasets, RWTH-Phoenix-2014T and How2Sign, are the most used datasets in SLP. The former includes German sign language sentences that can be used for text-to-sign language translation. The later is a recently proposed multi-modal dataset used for speech-to-sign language translation. Though RWTH-PHOENIX-Weather 2014T \cite{Camgoz-2018} and How2Sign \cite{Duarte-dataset} provided the appropriate SLP evaluation benchmarks, they are not enough for the generalization of the SLP models. Furthermore, these datasets only include German and American sentences. Translating from the spoken language to a large diversity of sign languages is a major challenge for the Deaf community.\\
\textbf{Applications:} American Sign Language (ASL), as the largest sign language community in the World, is a most-used sign language in the developed applications for SLP. Since Deaf people cannot read or write the spoken language, they need some tools for communication with the other people in society. Furthermore, many interesting and useful applications in Internet are not accessible for the Deaf community. To tackle these challenges, some projects have been proposed aiming to develop such tools. While these applications successfully made a bridge between Deaf and hearing communities, we are still far from having applications involving a large vocabularies/sentences from complex real-world scenarios. One of the main challenges for these applications is a license right for usage. Only some of these applications are freely available. Another challenge is regarding the application domain. Most of these applications have been developed for very specific domains. Improving the amount of available data and its quality can benefit the creation of these needed applications.\\  
\textbf{Proposed models:} The proposed works in SLP can be categorized into five categories: Avatar approaches, NMT approaches, MG approaches, Conditional image/video generation approaches, and other approaches. Table \ref{Table 2 } shows a summary of state-of-the-art deep SLP models. Some samples of the generated videos and gloss annotations are shown in Figure \ref{Fig 5}, \ref{Fig 6}, and \ref{Fig 7}. Using the data collected from motion capture, avatars can be more usable and acceptable for reviewers. Avatars achieve highly realistic results but the results are restricted to a small set of phrases. This comes from the cost of the data collection and annotation. Furthermore, avatar data is not a scalable solution and needs expert knowledge to be inspected and polished. To cope with these problems and improve performance, deep learning-based models are used.\\
While NMT-based methods achieved significant results in translation tasks, some major challenges need to be solved. Domain adaptation is the first challenge in this area. Since the translation in different domains need different styles and requirements, it is a crucial requirement in developing machine translation systems targeted at a specific use case. The second challenge is regarding the amount of available training data. Especially in deep learning-based models, increasing the amount of data can lead to better results. Another challenge is regarding to the uncommon words. The translation models perform poorly on these words. Words alignment and adjusting the beam search parameters are other challenges in NMT-based models.\\
Although MG can generate plausible and controllable motion through a database of motion capture, it faces some challenges. The first challenge is regarding limited access to data. To show the model potential with a truly diverse set of actions, a large set of data is necessary. Scalability and computational complexity of the graph to select the best transitions are other challenges in MG. Furthermore, since the number of edges leaving a node increases with the graph size, the branching factor in the search algorithm will increase as well. To automatically adjusting the graph configuration and relying on the training data, instead of user interference, Graph Convolutional Network (GCN) could be used along with some refining algorithms to adopt the graph structure monotonically.\\
While GANs has recently achieved remarkable results for image/video generation, there exist major challenges in the training of GANs. Mode collapse, non-convergence and instability, suitable objective function, and optimization algorithm are some of these challenges. However, several suggestions have been recently proposed to address the better design and optimization of GANs. Appropriate design of network architecture, proper objective functions, and optimization algorithms are some of the proposed techniques to improve the performance of GAN-based models. Finally, the challenge of the model complexity still remains for hybrid models.\\
\textbf{Limitations:} In this survey, we presented recent advances in SLP and related areas using deep learning. While successful results have been achieved in SLP by recent deep learning-based models, there are some limitations that need to be addressed. A main challenge is regarding the Multi-Signer (MS) generation that is necessary for providing a real-world communication in the Deaf community. To this end, we need to produce multiple signers of different appearance and configuration. Another limitation is the possibility of high resolution and photo-realistic continuous sign language videos. Most of the proposed models in SLP can only generate low resolution sign samples. Conditioning on human keypoints extracted from training data can decrease the parameter complexity of the model and assist to produce a high resolution video sign. However, avatar-based models can successfully generate high resolution video samples, though they are complex and expensive. In addition, pruning algorithms of MG need to be improved by including additional features of sign language, such as duration and speed of motion.\\ 
\textbf{Future directions:} While recent models in SLP presented promising results relying on deep learning capabilities, there is still much room for improvement. Considering the discriminative power of self-attention, learning to fuse multiple input modalities to benefit from multi-channel information, learning structured spatio-temporal patterns (such as Graph Neural Networks models), and employing domain-specific prior knowledge on sign language are some possible future directions in this area. Furthermore, there are some exciting assistive technologies for Deaf and hearing-impaired people. A brief introduction to these technologies can get an insight to the researchers in SLP and also make a bridge between them and the corresponding technology requirements. These technologies fall into three device categories: hearing technology, alerting devices, and communication support technology. For example, let imagine a technology that assists a Deaf person go through a musical experience translated into another sensory modality. While the recent advances in SLP are promising, more endeavor is indispensable to provide a fast processing model in an uncontrolled environment considering rapid hand motions. It is clear that technology standardization and full interoperability among devices and platforms are prerequisites to having real-life communication between the hearing and hearing-impaired communities.

{\small
\bibliographystyle{ieee_fullname}
\bibliography{egbib}

\begin{thebibliography}{10}\itemsep=-1pt

\bibitem{Arikan}
Okan Arikan and D.A. Forsyth.
\newblock Interactive motion generation from examples.
\newblock {\em In Proceedings of the 29th annual conference on computer
  graphics and interactive techniques, SIGGRAPH ’02}, pages 483--490, 2002.

\bibitem{china}
Start ASL.
\newblock Chinese sign language.
\newblock {\em https://www.startasl.com/chinese-sign-language/}, 2021.

\bibitem{spain}
Start ASL.
\newblock Spanish sign language.
\newblock {\em https://www.startasl.com/spanish-sign-language-ssl/}, 2021.

\bibitem{Athitsos}
Vassilis Athitsos, Carol Neidle, Stan Sclaroff, Joan Nash, Alexandra Stefan,
  Quan Yuan, and Ashwin Thangali.
\newblock The american sign language lexicon video dataset.
\newblock {\em CVPR}, pages 1--8, 2018.

\bibitem{Bachmann}
Daniel Bachmann, Frank Weichert, and Gerhard Rinkenauer.
\newblock Review of three-dimensional human-computer interaction with focus on
  the leap motion controller.
\newblock {\em Sensors}, 2018.

\bibitem{Bahdanau}
Dzmitry Bahdanau, KyungHyun Cho, and Yoshua Bengio.
\newblock Neural machine translation by jointly learning to align and
  translate.
\newblock {\em ICLR}, 2015.

\bibitem{Bangham}
J.~A. Bangham, S.~J. Cox, R. Elliott, J.~R.~W. Glauert, I. Marshall, S. Rankov,
  and M. Wells.
\newblock Virtual signing: Capture, animation, storage and transmission – an
  overview of the visicast project.
\newblock {\em Speech and Language Processing for Disabled and Elderly People},
  2000.

\bibitem{Bungeroth}
Jan Bungeroth, Daniel Stein, Philippe Dreuw, Hermann Ney, Sara Morrissey, Andy
  Way, and Lynette van Zijl.
\newblock The atis sign language corpus.
\newblock {\em 6th International Conference on Language Resources and
  Evaluation}, 2008.

\bibitem{Butt}
A.~H. Butt, E. Rovini, C. Dolciotti, G.~De Petris, P. Bongioanni, M.~C.
  Carboncini, and F. Cavallo.
\newblock Objective and automatic classification of parkinson disease with leap
  motion controller.
\newblock {\em BioMed Eng OnLine}, 17, 2018.

\bibitem{cai}
Su Cai, Gaoxia Zhu, Ying-Tien Wu, Enrui Liu, and Xiaoyi Hu.
\newblock A case study of gesture-based games in enhancing the fine motor
  skills and recognition.
\newblock {\em Interactive Learning Environments}, 26, 2018.

\bibitem{Camgoz-2018}
Necati~Cihan Camgoz, Simon Hadfield, Oscar Koller, Hermann Ney, and Richard
  Bowden.
\newblock Neural sign language translation.
\newblock {\em CVPR}, 2018.

\bibitem{Camgoz}
Necati~Cihan Camgoz, Oscar Koller, Simon Hadfield, and Richard Bowden.
\newblock Multi-channel transformers for multi-articulatory sign language
  translation.
\newblock {\em ECCVW}, 2020.

\bibitem{Necati2018}
Necati~Cihan Camgöz, Simon Hadfield, Oscar Koller, Hermann Ney, and Richard
  Bowden.
\newblock Rwth-phoenix-weather 2014 t: Parallel corpus of sign language video,
  gloss and translation.
\newblock {\em CVPR, Salt Lake City, UT}, 2018.

\bibitem{Caselli}
Naomi~K. Caselli, Zed~Sevcikova Sehyr, Ariel~M. Cohen-Goldberg, and Karen
  Emmorey.
\newblock Asl-lex: A lexical database for asl.
\newblock {\em Behavior Research Methods}, 49:784–801, 2017.

\bibitem{Chen}
Qifeng Chen and Vladlen Koltun.
\newblock Photographic image synthesis with cascaded refinement networks.
\newblock {\em ICCV}, pages 1511--1520, 2017.

\bibitem{Cohen}
Miri~Weiss Cohen, Israel Voldman, Daniele Regazzoni, and Andrea Vitali.
\newblock Hand rehabilitation via gesture recognition using leap motion
  controller.
\newblock {\em 11th International Conference on Human System Interaction (HSI),
  Gdansk, Poland}, 2018.

\bibitem{Cox}
Stephen Cox, Mike Lincoln, Judy Tryggvason, Melanie Nakisa, Mark Wells, Marcus
  Tutt, and Sanja Abbott.
\newblock Tessa, a system to aid communication with deaf people.
\newblock {\em In Proceedings of the 5th international ACM conference on
  assistive technologies}, pages 205--212, 2002.

\bibitem{Darabkh}
Khalid~A. Darabkh, Farah~H. Alturk, and Saadeh~Z. Sweidan.
\newblock Vrcdea‐tcs: 3d virtual reality cooperative drawing educational
  application with textual chatting system.
\newblock {\em Comput Appl Eng Educ}, 26:1677--1698, 2018.

\bibitem{Dawes}
Felix Dawes, Jaques Penders, and Giuseppe Carbone.
\newblock Remote control of a robotic hand using a leap sensor.
\newblock {\em The international conference of IFToMM ITALY}, pages 332--341,
  2018.

\bibitem{Duarte-dataset}
Amanda Duarte.
\newblock Cross-modal neural sign language translation.
\newblock {\em The 27th ACM International Conference}, 2019.

\bibitem{Duarte}
Amanda Duarte, Shruti Palaskar, Deepti Ghadiyaram, Kenneth DeHaan, Florian
  Metze, Jordi Torres, and Xavier GiroiNieto1.
\newblock How2sign: A large-scale multimodal dataset for continuous american
  sign language.
\newblock {\em Sign Language Recognition, Translation, and Production
  workshop}, 2020.

\bibitem{Ebling-2015}
Sarah Ebling and Matt Huenerfauth.
\newblock Bridging the gap between sign language machine translation and sign
  language animation using sequence classification.
\newblock {\em Proceedings of SLPAT 2015: 6th Workshop on Speech and Language
  Processing for Assistive Technologies}, pages 2--9, 2015.

\bibitem{EblingJohn-2013}
Sarah EblingJohn and GlauertJohn Glauert.
\newblock Exploiting the full potential of jasigning to build an avatar signing
  train announcements.
\newblock {\em In 3rd International symposium on sign language translation and
  avatar technology}, pages 1--9, 2013.

\bibitem{Efthimiou}
Eleni Efthimiou, Stavroula-Evita Fotinea, Thomas Hanke, John Glauert, Richard
  Bowden, Annelies Braffort, Christophe Collet, Petros Maragos, and François
  Lefebvre-Albaret.
\newblock The dicta-sign wiki: Enabling web communication for the deaf.
\newblock {\em International Conference on Computers for Handicapped Persons
  (ICCHP)}, pages 205--212, 2012.

\bibitem{argentina}
Ethnologue.
\newblock Argentine sign language.
\newblock {\em https://www.ethnologue.com/language/aed}, 2021.

\bibitem{Greek}
Ethnologue.
\newblock Greek sign language.
\newblock {\em https://www.ethnologue.com/language/gss}, 2021.

\bibitem{Iran}
Ethnologue.
\newblock Persian sign language.
\newblock {\em https://www.ethnologue.com/language/psc}, 2021.

\bibitem{Forster}
Jens Forster, Christoph Schmidt, Thomas Hoyoux, Oscar Koller, Uwe Zelle, Justus
  Piater, and Hermann Ney.
\newblock Rwth-phoenix-weather: A large vocabulary sign language recognition
  and translation corpus.
\newblock {\em Proceedings of the Eighth International Conference on Language
  Resources and Evaluation (LREC12), Istanbul, Turkey}, page 3785–3789, 2012.

\bibitem{Ghanem}
Sakher Ghanem, Christopher Conly, and Vassilis Athitsos.
\newblock A survey on sign language recognition using smartphones.
\newblock {\em Proceedings of the 10th international conference on pervasive
  technologies related to assistive environments, Island of Rhodes Greece},
  2017.

\bibitem{Gibet}
Sylvie Gibet, François Lefebvre-Albaret, Ludovic Hamon, Rémi Brun, and Ahmed
  Turki.
\newblock Interactive editing in french sign language dedicated to virtual
  signers: Requirements and challenges.
\newblock {\em Universal Access in the Information Society}, 15:525--539, 2016.

\bibitem{GAN}
Ian~J. Goodfellow, Jean Pouget-Abadie, Mehdi Mirza, Bing Xu, David
  Warde-Farley, Sherjil Ozair, Aaron Courville, and Yoshua Bengio.
\newblock Generative adversarial nets.
\newblock {\em Advances in Neural Information Processing Systems}, page
  2672–2680, 2014.

\bibitem{Gregor}
Karol Gregor, Ivo Danihelka, Alex Graves, Danilo Rezende, and Daan Wierstra.
\newblock Draw: A recurrent neural network for image generation.
\newblock {\em Proceedings of Machine Learning Research}, 2015.

\bibitem{VHG}
Virtual~Humans Group.
\newblock Virtual humans research for sign language animation.
\newblock {\em School of Computing Sciences, UEA Norwich, UK}, 2017.

\bibitem{Guo-2018}
Dan Guo, Wengang Zhou, Houqiang Li, and Meng Wang.
\newblock Hierarchical lstm for sign language translation.
\newblock {\em The Thirty-Second AAAI Conference on Artificial Intelligence
  (AAAI-18)}, 2018.

\bibitem{Germany}
Thomas Hanke.
\newblock German sign language.
\newblock {\em
  https://www.awhamburg.de/en/research/long-term-scientific-projects/dictionary-german-sign-language.html},
  2021.

\bibitem{Hutchins}
John Hutchins.
\newblock History of machine translation.
\newblock {\em http://psychotransling.ucoz.com/-ld/0/11-Hutchins-survey.pdf},
  2005.

\bibitem{Jemni}
Mohammed Jemni, Oussama~El Ghoul, Mehrez Boulares, Nour~Ben Yahia, Kabil
  Jaballah, Achraf Othman, and Monoem Youneb.
\newblock Websign.
\newblock {\em http://www.latice.rnu.tn/websign/}, 2020.

\bibitem{Kalchbrenner}
Nal Kalchbrenner, Lasse Espeholt, Karen Simonyan, Aaron van~den Oord, Alex
  Graves, and Koray Kavukcuoglu.
\newblock Neural machine translation in linear time.
\newblock {\em arXiv:1610.10099}, 2016.

\bibitem{Karras}
Tero Karras, Samuli Laine, and Timo Aila.
\newblock A style-based generator architecture for generative adversarial
  networks.
\newblock {\em CVPR}, 2019.

\bibitem{Kataoka}
Yuusuke Kataoka, Takashi Matsubara, and Kuniaki Uehara.
\newblock Image generation using adversarial networks and attention mechanism.
\newblock {\em IEEE/ACIS 15th International Conference on Computer and
  Information Science (ICIS)}, 2016.

\bibitem{Kennaway}
Richard Kennaway.
\newblock Avatar-independent scripting for real-time gesture animation.
\newblock {\em Procedural animation of sign language, arXiv:1502.02961}, 2013.

\bibitem{VAE}
Diederik~P Kingma and Max Welling.
\newblock Auto-encoding variational bayes.
\newblock {\em ICLR}, 2014.

\bibitem{Kipp-Avatar}
Michael Kipp, Alexis Heloir, and Quan Nguyen.
\newblock Sign language avatars: Animation and comprehensibility.
\newblock {\em International Workshop on Intelligent Virtual Agents}, pages
  113--126, 2011.

\bibitem{Ko}
Sang-Ki Ko, Chang~Jo Kim, Hyedong Jung, and Choongsang Cho.
\newblock Neural sign language translation based on human keypoint estimation.
\newblock {\em Applied Sciences}, 9, 2019.

\bibitem{Kovar}
Lucas Kovar, Michael Gleicher, and Frédéric Pighin.
\newblock Motion graphs.
\newblock {\em SIGGRAPH '02: Proceedings of the 29th annual conference on
  Computer graphics and interactive techniques}, pages 473--483, 2002.

\bibitem{Lee}
Jehee Lee and Sung~Yong Shin.
\newblock A hierarchical approach to interactive motion editing for human-like
  figures.
\newblock {\em SIGGRAPH '99: Proceedings of the 26th annual conference on
  Computer graphics and interactive techniques}, pages 39--48, 1999.

\bibitem{Luong}
Minh-Thang Luong, Hieu Pham, and Christopher~D. Manning.
\newblock Effective approaches to attention-based neural machine translation.
\newblock {\em arXiv:1508.04025}, 2015.

\bibitem{Ma-2017}
Liqian Ma, Xu Jia, Qianru Sun, Bernt Schiele, Tinne Tuytelaars, and Luc~Van
  Gool.
\newblock Pose guided person image generation.
\newblock {\em NIPS}, 2017.

\bibitem{Rastgoo-cnn}
Nezam Majidi, Kourosh Kiani, and Razieh Rastgoo.
\newblock A deep model for super-resolution enhancement from a single image.
\newblock {\em Journal of AI and Data Mining}, 8:451--460, 2020.

\bibitem{Matthes}
Silke Matthes, Thomas Hanke, Anja Regen, Jakob Storz, Satu Worseck, Eleni
  Efthimiou, Athanasia-Lida Dimou, Annelies Braffort, John Glauert, and Eva
  Safar.
\newblock Dicta-sign–building a multilingual sign language corpus.
\newblock {\em In 5th LREC. Istanbul}, 2012.

\bibitem{McDonald}
John McDonald, Rosalee Wolfe, Jerry Schnepp, Julie Hochgesang, Diana~Gorman
  Jamrozik, Marie Stumbo, Larwan Berke, Melissa Bialek, and Farah Thomas.
\newblock An automated technique for realtime production of lifelike animations
  of american sign language.
\newblock {\em Universal Access in the Information Society}, 15:551--566, 2016.

\bibitem{Morando}
Matteo Morando, Serena Ponte, Elisa Ferrara, and Silvana Dellepiane.
\newblock Definition of motion and biophysical indicators for home-based
  rehabilitation through serious games.
\newblock {\em Information}, 9, 2018.

\bibitem{Mori}
Masahiro Mori, Karl~F. MacDorman, and Norri Kageki.
\newblock The uncanny valley [from the field].
\newblock {\em IEEE Robotics and Automation Magazine}, 19:98--100, 2012.

\bibitem{Othman}
Achraf Othman and Mohamed Jemni.
\newblock Statistical sign language machine translation: from english written
  text to american sign language gloss.
\newblock {\em IJCSI International Journal of Computer Science}, 8:65--73,
  2011.

\bibitem{Korean}
Owlcation.
\newblock Korean sign language.
\newblock {\em https://owlcation.com/humanities/Korean-Sign-Language}, 2021.

\bibitem{Prillwitz}
Siegmund Prillwitz.
\newblock Hamnosys. version 2.0. hamburg notation system for sign languages. an
  introductory guide.
\newblock {\em Hamburg Signum Press}, 1989.

\bibitem{Rastgoo-rbm}
Razieh Rastgoo, Kourosh Kiani, and Sergio Escalera.
\newblock Multi-modal deep hand sign language recognition in still images using
  restricted boltzmann machine.
\newblock {\em Entropy}, 20, 2018.

\bibitem{Rastgoo-multiview}
Razieh Rastgoo, Kourosh Kiani, and Sergio Escalera.
\newblock Hand sign language recognition using multi-view hand skeleton.
\newblock {\em Expert Systems With Applications}, 150, 2020.

\bibitem{Rastgoo-video}
Razieh Rastgoo, Kourosh Kiani, and Sergio Escalera.
\newblock Video-based isolated hand sign language recognition using a deep
  cascaded model.
\newblock {\em Multimedia Tools And Applications}, 79:22965–22987, 2020.

\bibitem{Rastgoo-pose-aware}
Razieh Rastgoo, Kourosh Kiani, and Sergio Escalera.
\newblock Hand pose aware multimodal isolated sign language recognition.
\newblock {\em Multimedia Tools And Applications}, 80:127–163, 2021.

\bibitem{Rastgoo-svd}
Razieh Rastgoo, Kourosh Kiani, and Sergio Escalera.
\newblock Real-time isolated hand sign language recognition using deep networks
  and svd.
\newblock {\em Journal of Ambient Intelligence and Humanized Computing}, 2021.

\bibitem{RastgooSurvey}
Razieh Rastgoo, Kourosh Kiani, and Sergio Escalera.
\newblock Sign language recognition: A deep survey.
\newblock {\em Expert Systems With Application}, 164:113794, 2021.

\bibitem{Roccetti}
Marco Roccetti, Gustavo Marfia, and Angelo Semeraro.
\newblock Playing into the wild: A gesture-based interface for gaming in public
  spaces.
\newblock {\em Journal of Visual Communication and Image Representation},
  23:426--440, 2012.

\bibitem{poland}
Salto.
\newblock Polish sign language.
\newblock {\em
  https://www.salto-youth.net/tools/otlas-partner-finding/organisation/association-of-polish-sign-language-interpreters.2561/},
  2021.

\bibitem{SaundersBMVC}
Ben Saunders, Necati~Cihan Camgoz, and Richard Bowden.
\newblock Adversarial training for multi-channel sign language production.
\newblock {\em BMVC}, 2020.

\bibitem{Saunders}
Ben Saunders, Necati~Cihan Camgoz, and Richard Bowden.
\newblock Progressive transformers for end-to-end sign language production.
\newblock {\em ECCV}, pages 687--705, 2020.

\bibitem{Saunders-BMVC}
Ben Saunders, Necati~Cihan Camgöz, and Richard Bowden.
\newblock Adversarial training for multi-channel sign language production.
\newblock {\em BMVC}, 2020.

\bibitem{Saunders-arxiv}
Ben Saunders, Necati~Cihan Camgöz, and Richard Bowden.
\newblock Everybody sign now: Translating spoken language to photo realistic
  sign language video.
\newblock {\em arXiv:2011.09846}, 2020.

\bibitem{See}
Abigail See and Matthew Lamm.
\newblock Machine translation, sequence-to-sequence and attention.
\newblock {\em
  https://web.stanford.edu/class/cs224n/slides/cs224n-2020-lecture08-nmt.pdf},
  2021.

\bibitem{Siarohin}
Aliaksandr Siarohin, Enver Sangineto, Stephane Lathuiliere, and Nicu Sebe.
\newblock Deformable gans for pose-based human image generation.
\newblock {\em CVPR}, 2018.

\bibitem{Stoll-sign}
Stephanie Stoll, Necati~Cihan Camgoz, Simon Hadfield, and Richard Bowden.
\newblock Sign language production using neural machine translation and
  generative adversarial networks.
\newblock {\em BMVC}, 2018.

\bibitem{stoll2020}
Stephanie Stoll, Necati~Cihan Camgoz, Simon Hadfield, and Richard Bowden.
\newblock Text2sign: Towards sign language production using neural machine
  translation and generative adversarial networks.
\newblock {\em International Journal of Computer Vision}, 128:891--908, 2020.

\bibitem{Sutskever}
Ilya Sutskever, Oriol Vinyals, and Quoc~V. Le.
\newblock Sequence to sequence learning with neural networks.
\newblock {\em NIPS}, 2014.

\bibitem{Tornay}
Sandrine Tornay, Necati~Cihan Camgoz, Richard Bowden, and Magimai Doss.
\newblock A phonology-based approach for isolated sign production assessment in
  sign language.
\newblock {\em ICMI '20 Companion: Companion Publication of the 2020
  International Conference on Multimodal Interaction}, 2020.

\bibitem{Vaitkev}
Aurelijus Vaitkevičius, Mantas Taroza, Tomas Blažauskas, Robertas
  Damaševičius, Rytis Maskeliūnas, and Marcin Woźniak.
\newblock Recognition of american sign language gestures in a virtual reality
  using leap motion.
\newblock {\em Appl. Sci.}, 9, 2019.

\bibitem{Oord-pixelRNN}
Aaron van~den Oord, Nal Kalchbrenner, and Koray Kavukcuoglu.
\newblock Pixel recurrent neural networks.
\newblock {\em Proceedings of The 33rd International Conference on Machine
  Learning}, pages 1747--1756, 2016.

\bibitem{Oord}
Aaron van~den Oord, Nal Kalchbrenner, Oriol Vinyals, Lasse Espeholt, Alex
  Graves, and Koray Kavukcuoglu.
\newblock Conditional image generation with pixelcnn decoders.
\newblock {\em NIPS}, 2016.

\bibitem{Vasani}
Neel Vasani, Pratik~Autee an, and Samip Kalyani;~Ruhina Karani.
\newblock Generation of indian sign language by sentence processing and
  generative adversarial networks.
\newblock {\em International Conference on Intelligent Sustainable Systems
  (ICISS)}, 2020.

\bibitem{Vaswani}
Ashish Vaswani, Noam Shazeer, Niki Parmar, Jakob Uszkoreit, Llion Jones,
  Aidan~N. Gomez, Łukasz Kaiser, and Illia Polosukhin.
\newblock Attention is all you need.
\newblock {\em NIPS}, 2017.

\bibitem{asl}
William~G. Vicars.
\newblock American sign language.
\newblock {\em http://www.lifeprint.com/}, 2021.

\bibitem{who1}
{WHO: World Health Organization}.
\newblock Deafness and hearing loss.
\newblock {\em http://www.who.int/mediacentre/factsheets/fs300/en/}, 2021.

\bibitem{CVAE}
Xinchen Yan, Jimei Yang, Kihyuk Sohn, and Honglak Lee.
\newblock Attribute2image: Conditional image generation from visual attributes.
\newblock {\em ECCV}, page 776–791, 2016.

\bibitem{yang}
Hee-Deok Yang.
\newblock Sign language recognition with the kinect sensor based on conditional
  random fields.
\newblock {\em Sensors}, 15:135--147, 2014.

\bibitem{Zelinka}
Jan Zelinka and Jakub Kanis.
\newblock Neural sign language synthesis: Words are our glosses.
\newblock {\em WACV}, pages 3395--3403, 2020.

\bibitem{Zwitserlood}
Inge Zwitserlood, Margriet Verlinden, Johan Ros, and Sanny van~der Schoot.
\newblock Synthetic signing for the deaf: esign.
\newblock {\em http://www.visicast.cmp.uea.ac.uk}, pages 1--6, 2005.

\end{thebibliography}
}
\end{document}